# Semantically Cohesive Word Grouping in Indian Languages

**N J Karthika[1], Adyasha Patra[1], Nagasai Saketh Naidu[1],
Arnab Bhattacharya[2], Ganesh Ramakrishnan[1], Chaitali Dangarikar[2]**,
[1]Department of CSE, IIT Bombay, [2]Department of CSE, IIT Kanpur

## Abstract

Indian languages are inflectional and agglutinative and typically follow clause-free word order. The structure of sentences across most major Indian languages are similar when their dependency parse trees are considered. While some differences in the parsing structure occur due to peculiarities of a language or its preferred natural way of conveying meaning, several apparent differences are simply due to the granularity of representation of the smallest semantic unit of processing in a sentence. The semantic unit is typically a word, typographically separated by whitespaces. A single whitespace-separated word in one language may correspond to a group of words in another. Hence, grouping of words based on semantics helps unify the parsing structure of parallel sentences across languages and, in the process, morphology. In this work, we propose *word grouping* as a major preprocessing step for any computational or linguistic processing of sentences for Indian languages. Among Indian languages, since Hindi is one of the least agglutinative, we expect it to benefit the most from word grouping. Hence, in this paper, we focus on Hindi to study the effects of grouping. We perform quantitative assessment of our proposal with an intrinsic method that perturbs sentences by shuffling words as well as an extrinsic evaluation that verifies the importance of word grouping for the task of Machine Translation (MT) using decomposed prompting. We also qualitatively analyze certain aspects of the syntactic structure of sentences. Our experiments and analyses show that the proposed grouping technique brings uniformity in the syntactic structures, as well as aids underlying NLP tasks.

## 1 Introduction

The process of extracting meaningful phrases from sentences, known as chunking, is an important task in NLP. From a more granular level, the ability to identify semantic units of a sentence can be advantageous for a variety of NLP applications. In this paper, we discuss the importance of word grouping in a sentence, which together form a single, independent meaningful unit of the sentence.

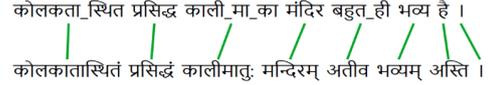

Figure 1: Alignment of parallel sentences in Hindi and Sanskrit, after word grouping.

Majority of Indian languages follow similar grammatical structure. The key changes in a syntactic structure like a dependency parse tree of a sentence, emerge mostly from differences in the number of whitespace-separated words[1] that represent a particular semantic concept. This variation occurs since we consider the words of a sentence as the basic units of processing. When we consider parallel sentences in various Indian languages, generally it is possible to obtain a non-overlapping word/phrase-level alignment. The major reason for not having a one-to-one mapping is the variation in the word count as discussed above. We find that grouping of words helps in a better alignment of Indian languages. Figure 1 displays a pair of parallel sentences in Hindi and Sanskrit. It shows how multiple words in one language correspond to a single word in another language.

Dangarikar et al. (2024) show that Hindi language exhibits a significant deviation from other major Indian languages. Data statistics provided by Gerz et al. (2018) using Polyglot Wikipedia also show a similar trend. The reason for such a deviation is that, among the Indian languages considered, Hindi is the least agglutinative in nature (Pimpale et al., 2014) and, at times, follow isolating features. Owing to such a deviation, we expect the word grouping effort to be more crucial and effective for Hindi and, hence, in this paper, we focus on word grouping for Hindi. For example, the words

---
[1]In the paper, usage of 'word' is for the whitespace-separated texts in any sentence.

जा रहा है (jā rahā hai[2]) in Hindi, corresponds to a single word hōgutiddāne in Kannada and yācchē in Bangla. Similarly, while Hindi tends to use "case-markers" such as kī, kē, etc., as separate words, highly agglutinated languages like Kannada and Malayalam use inflectional suffixes fused with the word. However, we emphasize that such differences are only at the typographic surface-level, and the *underlying semantic structure* of the languages is similar. Thus, (mōhana sē) in Hindi, (mōhanēna) in Sanskrit, and (mōhananimda) in Kannada, all have the same semantic structure: a nominal root, followed by a case-marker (in this example, the root is Mohan and the case-marker is instrumental). We can eliminate this dissimilarity by grouping the case-marker with the noun to form a single word group mōhana_sē.

Following are our contributions:
- We propose Indian-language-specific word grouping criteria to make the tokens semantically coherent.
- We propose a rule-based method to perform the word grouping task, by generating rules using a combination of data statistics and linguistically educated decisions.
- We show the importance of word grouping qualitatively and assess the importance quantitatively through intrinsic and extrinsic evaluation methods.

## 2 Related Works

### 2.1 Text Chunking vs word grouping

*Chunking* is an important preprocessing step for several NLP tasks, and is considered especially useful as a precursor for dependency parsing task (Abney, 2022). Other underlying tasks for which chunking plays an important role include Named Entity Recognition (Zhou and Su, 2002), information extraction (Dong et al., 2023), etc. Works on machine learning-based text chunking have been around for several decades (Church, 1989; Ramshaw and Marcus, 1999). Most of these works are based on English or related languages, and the most widely adopted granularity for chunking a sentence is phrases (noun phrases and verb phrases). Such a concept of phrasal chunks, though widely used, is not natural for Indian languages (Bharati et al., 1991). Bharati et al. (1991) showed the necessity for *Local Word Grouping (LWG)* in Indian languages. We extend the concept of LWG by defining a word group to be the smallest indivisible, semantically complete and meaningful unit of a sentence expressing a single linguistic function (known in Indian linguistic tradition as "ēkārthībhāva" and "sāmarthya"). The objective of word grouping is to make syntactic structures like dependency parse tree of a sentence, similar to that of parallel sentence in other Indian languages. This can further aid in cross-lingual NLP tasks.

### 2.2 Unfairness in Tokenization

Tokenization is a standard preprocessing step in NLP tasks, where a given input is broken into the smallest units for a system to process. Prior works (Petrov et al., 2024; Ahia et al., 2023) have shown the unfairness that arises in Language Models due to large variations in number of tokens associated with the same semantic content for different languages. In addition to this, we also see a language-dependent imbalance caused by the variation in the number of words used to convey the same concept in different languages (Dangarikar et al., 2024), which is not addressed in these works.

Generally, we consider the space-separated sequence of characters as a word. Considering the diversity of languages, the semantic information present in each word differs significantly. An isolating language like Chinese involves close to just one morpheme per word whereas, an agglutinated language like Malayalam has words that include multiple morphemes added sequentially, with various morphological information, such as gender, tense, person, etc., fused with the word in the form of affixes. Such a variation exist among multiple Indian languages too[3].

## 3 Methodology

Following are the basic rules followed in our proposal for grouping of words in accordance with Dangarikar et al. (2024).

- **Inflectional unity**: grouping nouns followed by post-positions, which are essentially the inflectional morphemes.
  *Example groups:* राम ने (rāma nē), हाथ से (hātha sē), बच्चों को (baccōm kō)
- **Derivational unity**: grouping verb and auxiliary verbs of a sentence, resulting in a single and complete action.
  *Example groups:* जा रहा है (jā rahā hai), कर दिया गया (kara diyā gayā)

---

[2]ISO15919 Indic Transliteration scheme

[3]We add statistics for the major languages in Appendix A

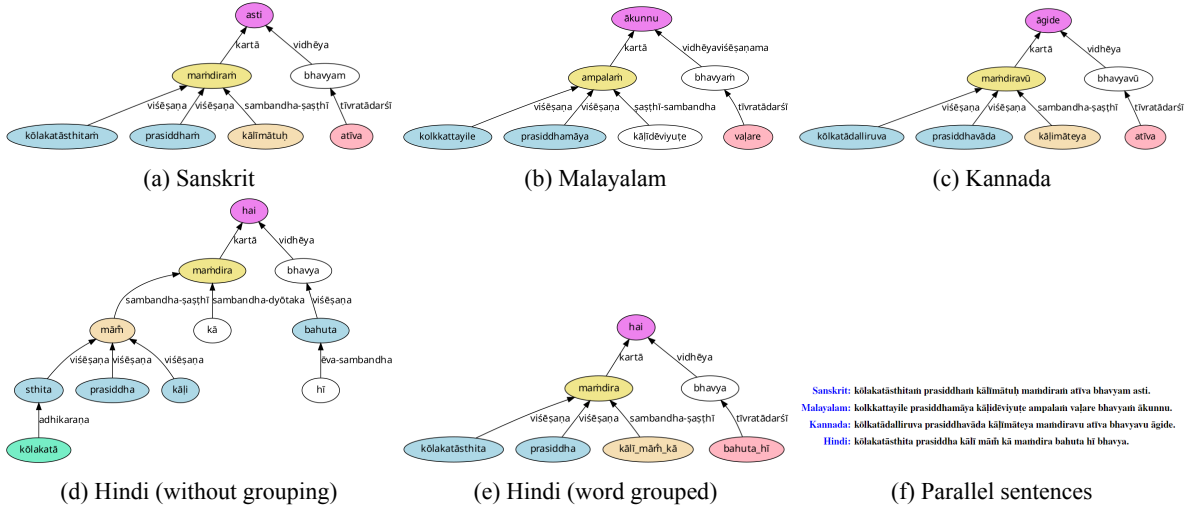

Figure 2: Dependency Parse Trees

- **Named entities**: A named entity (NE) with multiple words form a single group.
  *Example groups:* श्री ए पी जे अब्दुल कलाम (śrī ē pī jē abdula kalāma), अरुणाचल प्रदेश (aruṇācala pradēśa)

### 3.1 Similar Syntactic Structures

A dependency parse tree is a syntactic structural representation of a sentence, where the words or phrases form the nodes, and the edges show the dependencies between the nodes and their syntactic roles in the sentence.

Figure 2 shows dependency parse trees[4] of the sentences in Figure 2f, following (Dangarikar et al., 2024). Each word in the sentence forms a node in the tree. Figure 2e is the dependency parse tree for the Hindi sentence obtained after word grouping. Notice the similarity in the structure of the parse trees after grouping, while the tree structure of Hindi example was very different from others before word grouping, as seen in Figure 2d.

**Rule-Based Word Grouping**

In this section, we present the process followed to automatically generate word groups for Hindi data. The method used, though atypical in nature, generates good quality grouped data for Hindi.

We used (Kosaraju et al., 2012) data with kāraka-based dependency tags (Tandon et al., 2016), a widely used treebank dataset for Hindi, to statistically generate rules for word grouping.[5] The rules are generated from the dataset by finding the most frequent dependency relations between consecutive words in the sentence, along with the respective frequent POS tags of the tokens. We finalized the rules after verification by language-experts.

---
[4]Drawn using anvaya chitranam
[5]Generated rules are added in the Appendix A

Figure 3: (Top) words are randomly jumbled (Bottom) jumbled sentence with word-groups preserved

## 4 Experiments and Results

### 4.1 Sentence Perturbation

To justify the requirement for word grouping, we design an intrinsic evaluation method using sentence perturbation by jumbling words, a commonly adopted method to evaluate representational co-relatedness in sentences (Alleman et al., 2021; Sai et al., 2021). Our hypothesis is that word grouping allows sentences to preserve semantic roles/identities of its components, even on random shuffling.

A simple method of randomly jumbling words has a drawback when the sentences under consideration are long and complex, containing multiple clauses. Each clause may contain its own set of subject, object, verb, etc., and meanings may not be preserved despite the word groups being preserved when the words or groups are jumbled across multiple clauses. To address this issue, we experiment with a few different settings for grouping. (i) We perform random shuffling of the space-separated words vs inter-group shuffling, (ii) local shuffling of word units by fixing a window with a word count of 5 and only perform the intra-context shuffling, (iii) intra-context shuffling with a context size of 10 words, and (iv) a subset of the data containing less than 20 words per sentence, with the assumption that sentences with larger number of words are more likely to have multiple clauses. This sample contained 552 sentences.

| Setting | Grouped / Not | Languages | | | | | | | |
|---|---|---|---|---|---|---|---|---|---|
| | | Hindi | Kannada | Malayalam | Sanskrit | Tamil | Telugu | Bangla | Marathi |
| (i) | Ungrouped | 0.867 | 0.705 | 0.716 | 0.681 | 0.695 | 0.693 | 0.703 | 0.718 |
| | Grouped | **0.899** | **0.719** | **0.731** | **0.685** | **0.711** | **0.710** | **0.719** | **0.732** |
| (ii) | Ungrouped | 0.902 | 0.716 | 0.727 | **0.684** | 0.704 | 0.705 | 0.729 | 0.719 |
| | Grouped | **0.922** | **0.720** | **0.733** | 0.682 | **0.710** | **0.712** | **0.736** | **0.725** |
| (iii) | Ungrouped | 0.838 | 0.681 | 0.689 | **0.665** | 0.669 | 0.670 | 0.690 | 0.682 |
| | Grouped | **0.860** | **0.689** | **0.697** | 0.664 | **0.676** | **0.678** | **0.698** | **0.690** |
| (iv) | Ungrouped | 0.868 | 0.706 | 0.716 | 0.681 | 0.696 | 0.693 | 0.718 | 0.703 |
| | Grouped | **0.899** | **0.719** | **0.731** | **0.685** | **0.711** | **0.710** | **0.732** | **0.720** |

Table 1: Cosine similarity between shuffled Hindi sentences with the parallel sentences in other Indian languages. In every experiment, shuffling is done with and without the word-groups being preserved. (i) Shuffling entire sentence (ii) Local Shuffling within word window of size 5 (iii) Local Shuffling within word window of size 10 (iv) Data subset: Sentences with #words < 20.

| Languages | DecoMT w/o grouping | | DecoMT with grouping | |
|---|---|---|---|---|
| Source → Target | spBLEU | chrF++ | spBLEU | chrF++ |
| Hindi → Malayalam | 18.9 | 36.87 | **19.4** | **37.29** |
| Hindi → Kannada | 19.2 | 37.55 | **19.5** | **38.21** |
| Hindi → Sanskrit | 4.3 | 19.34 | **4.7** | **20.77** |
| Hindi → Bengali | 19.3 | 36.35 | **19.6** | **36.69** |
| Hindi → Marathi | 14.0 | 35.34 | **14.6** | **35.96** |

Table 2: Performance comparison of DecoMT (Few-Shot) preserving word groups, against the baseline that use fixed-length chunks.

We generate sentence embeddings for both sets of sentences using (Reimers, 2019) and cosine similarity to measure the similarity of the jumbled sentences with both the original unshuffled sentence and parallel sentences in 7 languages (Table 1). In most most of the cases, jumbled sentences with word groups preserved show higher similarity to original sentences. This establishes the significance of word grouping.

### 4.2 Decomposed Few-Shot Prompting

Puduppully et al. (2023) have shown improvements in MT task through few-shot prompting between related languages using decomposed prompts (DecoMT). They use mT5-XL(Xue, 2020) with 3.7B parameters as the base model for their experiments. DecoMT performs a combination of chunk-based translation and an iterative contextual translation and learns a combined loss. Input sentences are segmented by splitting them as fixed size word chunks. We observe that this method causes a meaningful semantic unit to be split across segments, which may ultimately reduce the translation quality, especially at the segment level. Considering word groups as a single unit prevents this split into different decompositions within a prompt. With this hypothesis, we perform experiments with DecoMT using FLORES-200 data (Costa-jussà et al., 2022), with the chunks as in (Puduppully et al., 2023), versus the chunks with word groups preserved. We perform the experiments to translate sentences from Hindi to 5 other languages. Table 2 shows consistent improvement on preserving word groups.

## 5 Conclusions

We propose the grouping of whitespace-separated words based on semantics, as a major preprocessing step for any computational and linguistic processing of sentences. Given the least agglutinative nature of Hindi compared to other Indian languages, we focus our experiments on Hindi, expecting it to benefit the most from grouping. We perform an intrinsic experiment, and an extrinsic experiment on MT. From both sets of experiment results, it is evident that a word group as a single semantic unit of a sentence provides a consistent improvement across experiments.

## Limitations

The process used for automatic word grouping is not straightforward. Though it generated a good quality word grouped data, it involves a dependence on another deep learning model. There is a need to have a more explicit method to generate grouping rules for automatic word grouping. For highly agglutinated languages, more than grouping, there may be a requirement to split a space-separated word into constituents, which we do not do in this work.

We intend to further simplify and facilitate the process of automatic word grouping in Hindi and also extend the grouping process (also splitting where necessary) to other languages.

# A Appendix

## A.1 Word count imbalances across languages

Generally, we consider the space-separated sequence of characters as a word. Considering the diversity of languages, the semantic information present in each word differs significantly. An isolating language like Chinese involves close to just one morpheme per word. In contrast, an agglutinated language like Malayalam has words that include multiple morphemes added sequentially, with various morphological information, such as gender, tense, person, etc., fused with the word in the form of inflectional affixes[6].

Figure 4 shows the total number of words in different Indian languages for the parallel sentences, representing the same content in FLORES-200 devtest data (Costa-jussà et al., 2022). The graph also contains some non-Indian languages to show the extent to which the number of words can vary across languages to convey the same information. Note that, the number of words in Jingpho is over $3.6\times$ the number of words in the corresponding parallel data in 'Shan'. In cross-lingual tasks involving languages exhibiting such variations, the associated models are responsible for implicitly learning the semantic unit-level mapping in addition to the underlying task, adding to the model complexity.

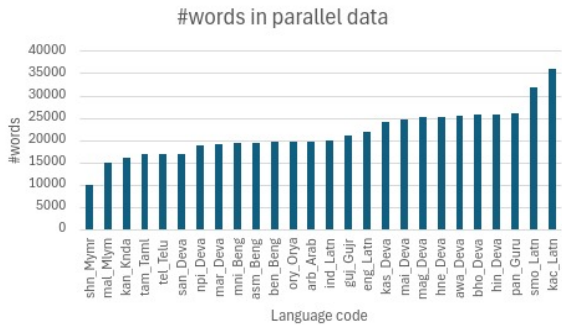

Figure 4: Number of space-separated words present in the parallel sentences of a subset of languages in FLORES-200 devtest data

Table 3 shows the word count of parallel sentences from a commonly used benchmark data for MT evaluation, *viz.,* FLORES-200 (Costa-jussà et al., 2022) devtest data. From the table, it is evident that, out of the six languages considered(Malayalam, Kannada, Sanskrit, Marathi, Bengali, and Hindi), Hindi exhibits a significant deviation from the rest. Data statistics provided by Gerz et al. (2018) using Polyglot Wikipedia, also show a similar trend. The reason for such a deviation is that among the Indian languages considered, Hindi is the least agglutinative in nature(Pimpale et al., 2014), and at times follow isolating features. In this paper, we particularly focus on grouping of words in Hindi, due to this distinctive feature that makes it to deviate the most from other languages. In light of such a deviation, we expect the grouping effort

---

[6]The words case-marker, vibhakti-marker, inflectional affixes and post-positions are used interchangeably in the paper

| Language | Total #words |
|---|---|
| Malayalam | 15001 |
| Kannada | 16577 |
| Sanskrit | 16992 |
| Marathi | 19046 |
| Bengali | 19585 |
| Hindi (without grouping) | 25643 |
| Hindi (grouped) | 18980 |

Table 3: Total number of whitespace separated words (or semantic units) in FLORES-200 devtest data. Hindi, when grouped, has a count similar to other Indian languages.

to be more crucial and effective for Hindi.

### A.2 Rules for Automatic Word Grouping

Table 4 shows the rules used to perform automatic word grouping for sentences. The rules are generated by a combination of treebank data statistics and feedback from linguists. The sentences are first input to trankit (Van Nguyen et al., 2021) tool to generate the corresponding values as shown in the table fields. This step is followed by the application of rules for grouping.

### A.3 Scores of Translation across different Sentence-Lengths

The DecoMT approach translates source sentences in sequential chunks, and we hypothesize that integrating word grouping will enhance translation adequacy, as it ensures that meaningful semantic units remain intact within chunks rather than being split across them. To investigate this, following a method similar to Puduppully et al. (2023), we categorize source sentences into buckets based on length, where each bucket's width corresponds to the standard deviation of the source sentence lengths. Buckets with fewer than 20 instances are merged with neighbouring ones. Figure 7 illustrates the relationship between source sentence length and chrF++ scores for translations from Hindi to Malayalam, Kannada, Sanskrit, Bengali, and Marathi. For all language pairs, we observe that DecoMT with word grouping consistently outperforms DecoMT in terms of chrF++ scores.

### A.4 Dataset and Evaluation Metrics

For both sets of experiments (Sentence Perturbation and DecoMT translation evaluation), we utilise a commonly used Benchmark dataset for Machine

| Dependent token | | Head | | Dependency relation | Examples |
|---|---|---|---|---|---|
| **Category** | **POS tag** | **Category** | **POS tag** | - | - |
| pn/n/v | NN/NNP/VM/PRP/PSP | psp | PSP | lwg_psp | हॉल(h)-में(d) ; जाने-के-लिए ; लगने-वाला |
| v | VAUX/VM | v | VAUX | lwg_vaux_cont | जाता-था ; रहता-है ; हुआ-है ; गए-है |
| v | VM | v | VAUX | lwg_vaux | बसाया-गया ; कहती-है ; बनवाया-था; बनी-हुई देती-है ; लिए-हुए |
| v | VM | avy | RP | lwg_rp | देखते-ही ; बड़ा-सा; स्थान-में-भी; भीड़ लगी-ही रहती है; कितने-ही; एक-ही |
| n | NN | v | JJ | pof | बंद-रहता; आनंद-उठा ; स्मृति-दिलाता ; अलग-होना ; कैद-होना |
| n/adj | NNPC/NNC | any | any | pof_cn | म.प्र.(d)-पर्यटन(h)-बोट(h)-क्लब(h); विमान-सेवा ; 17वीं-शताब्दी; रुद्र-प्रताप |
| n/adj | NN/NNP | any | any | pof | क्या नहीं-किया ; प्रवेश नही-मिलता |

Table 4: Rules used for Automatic Word Grouping

Translation Evaluation, FLORES-200 (Costa-jussà et al., 2022), specifically the dev-test data, which has 1012 sentences, parallelly available in 200 languages.

In the experiments discussed in Section 4.2, we perform the experiments to translate sentences from Hindi to five other languages *viz.,* Malayalam, Kannada, Sanskrit, Bengali and Marathi. We used BLEU and chrF++ scores to present the results. The specific signatures used for the metrics are

***BLEU Signature:*** nrefs:1| case:mixed| eff:no| tok:flores200| smooth:exp| version:2.4.2 and
***chrF++ Signature:*** nrefs:1| case:mixed| eff:yes| nc:6| nw:2| space:no| version:2.3.1.

*Note:* For MT experiments, we chose the three target languages, which are agglutinative in nature, with the intuition that the splitting at a sub-group level in Hindi sentences may cause decline in quality because of non-splittable words in these languages.

**Translate from Hindi to Bengali:**
Hindi: सोमवार को, स्टैनफ़ोर्ड यूनिवर्सिटी स्कूल
Bengali: সোমবারে স্ট্যানফোর্ড ইউনিভার্সিটি স্কুল
Hindi: ऑफ़ मेडिसिन के वैज्ञानिकों ने
Bengali: অফ মেডিসিনের বিজ্ঞানীরা
Hindi: एक नए डायग्नोस्टिक उपकरण के
Bengali: একটি নতুন ডায়াগনস্টিক উপকরণের
Hindi: आविष्कार की घोषणा की
Bengali: আবিষ্কারের ঘোষণা করেছেন
Hindi: कोशिकाओं को उनके प्रकार के
Bengali: কোষকে তাদের প্রকারের
Hindi: आधार पर छाँट सकने वाले,
Bengali: উপর ভিত্তি করে বাছাইযোগ্য,
Hindi: एक छोटी प्रिंट करने योग्य
Bengali: একটি ছোটো মুদ্রণযোগ্য
Hindi: चिप जिसे स्टैण्डर्ड इंकजेट प्रिंटर
Bengali: চিপ যেটা স্ট্যান্ডার্ড ইঙ্কজেট প্রিন্টারের
Hindi: का उपयोग करके लगभग
Bengali: ব্যবহার করে প্রায়
Hindi: एक अमेरिकी सेंट के लिए
Bengali: এক মার্কিন সেন্টের মধ্যে
Hindi: निर्मित किया जा सकता है।
Bengali: নির্মাণ করা যেতে পারে।

…3 more examples here

**Translate from Hindi to Bengali:**
Hindi: पायलट की पहचान स्क्वाड्रन लीडर
Bengali: পাইলটের পরিচয় স্কোয়াড্রন লিডার
Hindi: दिलोकृत पटावी के रूप में
Bengali: দিলোকৃত পাতাভির রূপে
Hindi: की गई।
Bengali: করা হয়েছে।

**Translate from Hindi to Bengali:**
Hindi: जीवित रहने की दर
Bengali: <mask>

Figure 5: DecoMT Prompt Template for Independent Translation with a Test Example: The template includes five sentences in the source (Hindi) and target (Bengali) languages divided into word chunks. The model receives a test example source chunk and a target language prompt with a □mask□ placeholder, aiming to predict the corresponding target chunk

**Translate from Hindi to Bengali:**
Hindi: सोमवार को, स्टैनफ़ोर्ड यूनिवर्सिटी स्कूल ऑफ़ मेडिसिन के
Bengali: সোমবারে স্ট্যানফোর্ড ইউনিভার্সিটি স্কুল অফ মেডিসিনের
Hindi: वैज्ञानिकों ने एक नए डायग्नोस्टिक उपकरण के
Bengali: বিজ্ঞানীরা একটি নতুন ডায়াগনস্টিক উপকরণের
Hindi: आविष्कार की घोषणा की
Bengali: আবিষ্কারের ঘোষণা করেছেন
Hindi: कोशिकाओं को उनके प्रकार के आधार पर
Bengali: কোষকে তাদের প্রকারের উপর ভিত্তি করে
Hindi: छाँट सकने वाले, एक छोटी प्रिंट करने योग्य
Bengali: বাছাইযোগ্য, একটি ছোটো মুদ্রণযোগ্য
Hindi: चिप जिसे स्टैण्डर्ड इंकजेट प्रिंटर का उपयोग करके
Bengali: চিপ যেটা স্ট্যান্ডার্ড ইঙ্কজেট প্রিন্টারের ব্যবহার করে
Hindi: लगभग एक अमेरिकी सेंट के लिए
Bengali: প্রায় এক মার্কিন সেন্টের মধ্যে
Hindi: निर्मित किया जा सकता है।
Bengali: নির্মাণ করা যেতে পারে।

…3 more examples here

**Translate from Hindi to Bengali:**
Hindi: पायलट की पहचान स्क्वाड्रन लीडर
Bengali: পাইলটের পরিচয় স্কোয়াড্রন লিডার
Hindi: दिलोकृत पटावी के रूप में की गई।
Bengali: দিলোকৃত পাতাভির রূপে করা হয়েছে।

**Translate from Hindi to Bengali:**
Hindi: जीवित रहने की दर आधी हो सकती है।
Bengali: <mask>

Figure 6: Proposed Prompt Template for Independent Translation with a Test Example: The template includes five sentences in the source (Hindi) and target (Bengali) languages divided into word chunks according to the word groupings. The model receives a test example source chunk and a target language prompt with a □mask□ placeholder, aiming to predict the corresponding target chunk

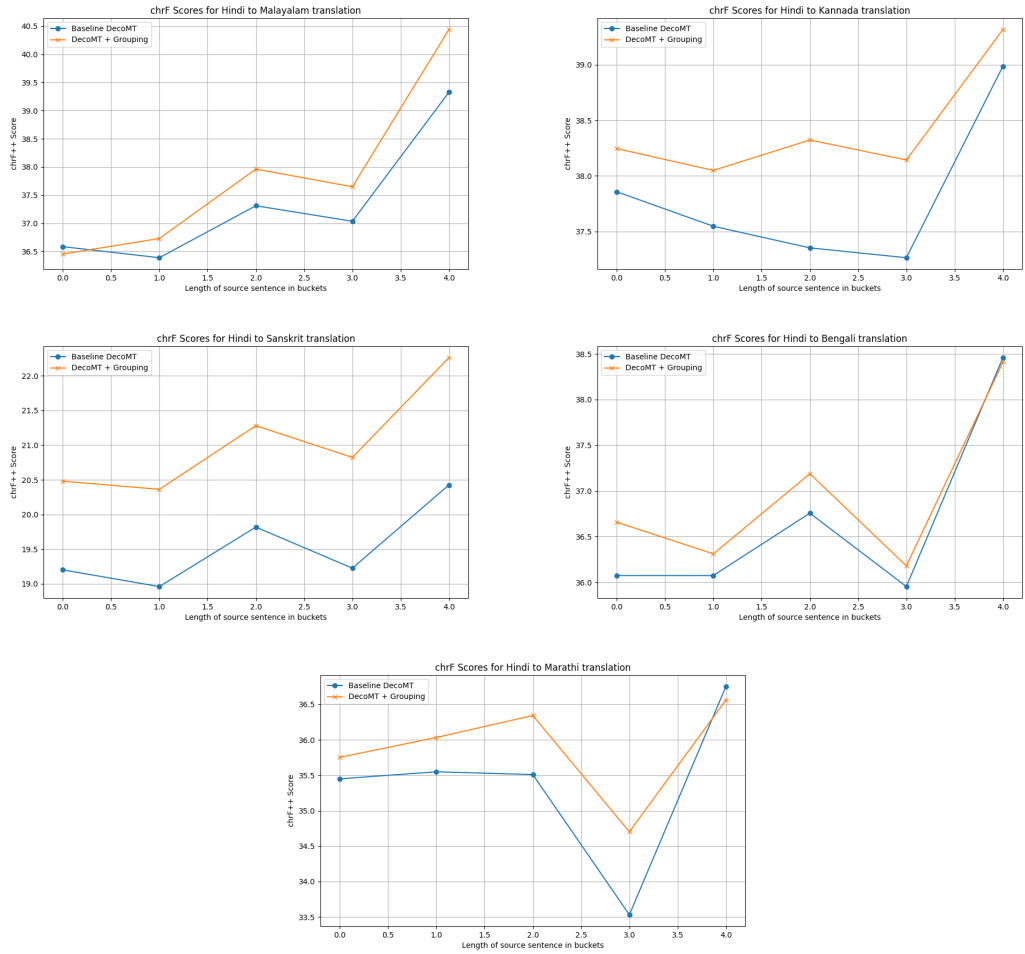

Figure 7: The plots show the relationship between source sentence length and chrF++ scores for translation from Hindi to Malayalam, Kannada, Sanskrit, Bengali, and Marathi. Lengths are bucketed, each equal to the source sentence lengths' standard deviation, with any bucket with less than 20 sentences merged with its neighbor. The data implies the chrF++ scores of DecoMT combined with our grouping outperform baseline DecoMT's performance.